\documentclass[11pt]{article}

\usepackage[preprint]{acl}

\usepackage{times}
\usepackage{latexsym}
\usepackage{comment}
\usepackage[T1]{fontenc}

\usepackage[utf8]{inputenc}

\usepackage{microtype}

\usepackage{inconsolata}

\usepackage{graphicx}

\usepackage{booktabs}
\usepackage{amsmath}
\usepackage{hyperref}
\usepackage{xcolor}

\usepackage{csquotes}

\usepackage{tabularx}

\usepackage{tikz}

\title{InterView-C: A Synchronized Multimodal Corpus of VR Avatar-Mediated Survey Interviews}

\author{
\begin{tabular}{cc}
\textbf{Patrick Schrottenbacher}\textsuperscript{1} &
\textbf{Leon Hammerla}\textnormal{\textsuperscript{1}}\\
\texttt{schrottenbacher@em.uni-frankfurt.de} &
\textnormal{\texttt{hammerla@em.uni-frankfurt.de}}
\end{tabular}
\\[1.2em]
\begin{tabular}{ccc}
\textbf{Lydia Kleine}\textsuperscript{2} &
\textbf{Doris Stingl}\textsuperscript{2} &
\textbf{Alexander Mehler}\textsuperscript{1} \\
\texttt{lydia.kleine@lifbi.de} &
\texttt{doris.stingl@lifbi.de} &
\texttt{mehler@em.uni-frankfurt.de}
\end{tabular}
\\[1.4em]
\hspace*{1.2cm}
\begin{minipage}{0.82\textwidth}
\centering
\textsuperscript{1} Goethe University,
Frankfurt am Main, Germany \\
\textsuperscript{2} Leibniz Institute for Educational Trajectories (LIfBi),
Bamberg, Germany
\end{minipage}
}

\begin{document}
\maketitle
\begin{abstract}
We present InterView-C, a German multimodal corpus of 27 survey interviews conducted entirely in virtual reality, with both interlocutors represented by avatars.
The corpus aligns spoken interaction with synchronized behavioral data, including gaze, head and body movement, facial behavior, hand and finger tracking.
Its reference transcripts and linguistic annotations provide a reliable interface between this multimodal spoken interaction and predominantly text-based NLP methods.
This interface is important because automatically transcribing speech can distort linguistically relevant information, while downstream models trained on existing resources may additionally face transfer challenges when applied to transcribed spoken data.
InterView-C therefore provides word-timed and manually post-edited verbatim transcripts for all 54 recordings, interview-item timings, questionnaire responses and negation cue and scope annotations for 1,422 sentences, 1,398 of them doubly annotated ($\alpha=0.87$ for cues; $\alpha=0.81$ for scopes).
We demonstrate both challenges empirically: nine open-weight ASR systems disproportionately misrecognize short closed answers and number words, while negation models trained on existing corpora show lower and highly variable performance on our transcribed interviews than a model trained on the InterView-C annotations.
InterView-C thus enables linguistic analyses of spoken interaction while retaining their alignment with rich multimodal behavior.
\end{abstract}

\section{Introduction}
Face-to-face interaction between two or more interlocutors is a complex system that extends well beyond the vocal channel.
Facial expressions, gaze and hand and arm gestures are among the many signals that accompany speech in face-to-face conversation~\citep{Vigliocco:et:al:2014,Holler:Levinson:2019}.
These signals can be analyzed separately, but each offers insight into only a thin slice of the interaction.
A holistic understanding of a conversation therefore requires capturing these interlinked modalities simultaneously.
This requires (a) recording the signals in a machine-readable, synchronized format and (b) methods to process and analyze these data.
As declining response rates and rising fieldwork costs pose challenges to survey research~\citep{Williams:Brick:2018,Luiten:et:al:2020}, new interview modes are being developed, ranging from computer-assisted web and live video interviews (CAWI, CALVI) to AI-assisted interviewing systems.
Besides reducing costs, these modes facilitate the recording of additional modalities, such as participants' audio and video (subject to consent and GDPR compliance), providing further insights into their behavior.
Access to these additional modalities allows further insights into participants' behaviors to be inferred, which is particularly valuable in the field of survey research.

These among others motivated the development of InterView~\citep{Schrottenbacher:et:al:2026:c}, a VR interview system built on Va.Si.Li-Lab~\citep{Mehler:et:al:2023} that has a particular focus on capturing and contextualizing these multimodal signals.
\usetikzlibrary{positioning,fit,calc,backgrounds,arrows.meta}
\pgfdeclarelayer{ivback}
\pgfsetlayers{ivback,background,main}

\definecolor{ivSession}{HTML}{0072B2}
\definecolor{ivTracking}{HTML}{009E73}
\definecolor{ivInteraction}{HTML}{E69F00}
\definecolor{ivAnnotation}{HTML}{E69F00}
\definecolor{ivLanguage}{HTML}{6A51A3}
\definecolor{ivInstrument}{HTML}{D55E00}
\definecolor{ivSelfReport}{HTML}{CC79A7}
\definecolor{ivEdge}{HTML}{707070}
\definecolor{ivInk}{HTML}{3A3A3A}

\newlength\ivCardW
\newlength\ivCardTW
\newcommand\ivSetCardWidth[1]{%
  \setlength\ivCardW{#1}\setlength\ivCardTW{\dimexpr#1-4.4mm\relax}}
\ivSetCardWidth{3.0cm}
\newcommand\ivTitleFont{\small\bfseries}
\newcommand\ivCountFont{\footnotesize}
\newcommand\ivAccent{black}

\newcommand\ivLayerRaw[2]{{\fontsize{5.6}{6.4}\selectfont\bfseries\color{#1}\MakeUppercase{#2}}}
\newcommand\ivLayer[2]{\ivLayerRaw{#1!80!black}{#2}}
\newcommand\ivTitle[1]{{\ivTitleFont\color{black!90}#1}}
\newcommand\ivCount[1]{{\ivCountFont\color{black!62}#1}}
\tikzset{
  iv accent/.code={\def\ivAccent{#1}},
  iv card/.style={draw=black!28, fill=white, line width=0.4pt, rounded corners=1.2pt,
    text width=\ivCardTW, minimum width=\ivCardW, inner xsep=2.2mm, inner ysep=1.0mm,
    align=left, font=\footnotesize,
    path picture={\fill[\ivAccent] (path picture bounding box.north west)
        rectangle ([xshift=1.7pt]path picture bounding box.south west);}},
  iv ghost/.style={iv card, draw=black!40, dash pattern=on 1.6pt off 1.2pt, fill=black!2},
  iv has/.style={-{Stealth[length=3.6pt,width=3pt]}, draw=ivEdge, line width=0.5pt},
  iv transcribed/.style={iv has, dash pattern=on 2pt off 1.4pt},
  iv flow/.style={-{Latex[length=4.2pt,width=3.6pt]}, draw=ivInk, line width=0.8pt},
  iv group/.style={draw=black!18, fill=black!3, rounded corners=2pt, inner sep=1.6mm},
  iv grouplabel/.style={font=\fontsize{5.6}{6.4}\selectfont\bfseries, text=black!58,
    anchor=south west, inner xsep=0.4pt, inner ysep=1.2pt},
  iv note/.style={font=\fontsize{6.8}{8}\selectfont, text=black!65, align=left, inner sep=0pt,
    execute at begin node={\hyphenpenalty=10000\relax\exhyphenpenalty=10000\relax}},
  iv sublabel/.style={font=\fontsize{6.6}{7.5}\selectfont\bfseries, text=black!70, inner sep=0pt, anchor=west},
  iv key/.style={font=\scriptsize, text=black!70, inner sep=0.6pt, anchor=west},
  iv num/.style={circle, fill=ivInk, text=white, inner sep=0pt, minimum size=3.6mm,
    font=\fontsize{6.5}{7}\selectfont\bfseries},
  iv badge/.style={rectangle, rounded corners=1.45mm, fill=black!50, text=white,
    minimum height=2.9mm, minimum width=2.9mm, inner xsep=1.3pt, inner ysep=0pt,
    font=\fontsize{5.4}{6}\selectfont\bfseries},
  iv step/.style={draw=black!35, fill=black!3, rounded corners=2pt, align=left,
    font=\scriptsize, execute at begin node={\hyphenpenalty=10000\relax\exhyphenpenalty=10000\relax}, text=black!80,
    inner xsep=1.6mm, inner ysep=1.3mm},
}

\newcommand\ivCard[6][]{%
  \node[iv card, iv accent=#3, #1] (#2)
    {\ivLayer{#3}{#4}\\[-0.35ex]\ivTitle{#5}\\[-0.35ex]\ivCount{#6}};}
\newcommand\ivGhost[5][]{%
  \node[iv ghost, iv accent=black!30, #1] (#2)
    {\ivLayerRaw{black!45}{#3}\\[-0.35ex]{\ivTitleFont\color{black!50}#4}\\[-0.35ex]{\ivCountFont\color{black!50}\itshape #5}};}
\newcommand\ivBadge[2]{\node[iv badge, anchor=north east] at ([shift={(-0.9mm,-0.9mm)}]#1.north east) {#2};}

\begin{figure*}[th]
\centering
\begin{tikzpicture}[node distance=2.2mm and 4.2mm, iv card/.append style={inner ysep=0.7mm}]
  \ivSetCardWidth{2.5cm}
  \renewcommand\ivTitleFont{\fontsize{7.5}{9}\selectfont\bfseries}
  \renewcommand\ivCountFont{\fontsize{7.5}{9}\selectfont}

  \ivCard{Experiment}{ivSession}{Session}{Experiment}{27}
  \ivCard[right=of Experiment]{Player}{ivSession}{Session}{Player}{54}
  \ivCard[right=7mm of Player]{Track}{ivTracking}{Per frame}{Tracking}{1,350,572 frames}
  \ivGhost[below=of Track]{Audio}{Media}{Audio}{withheld}
  \ivCard[below=of Audio]{Word}{ivLanguage}{Language}{Word}{53,475 (automatic)}
  \ivCard[at={(Player |- Audio)}]{QW}{ivInstrument}{Instrument}{QuestionWindow}{1,147}
  \ivCard[at={(Experiment |- Audio)}]{IItem}{ivInstrument}{Instrument}{InterviewItem}{56}
  \ivCard[at={(Experiment |- Word)}]{SAns}{ivSelfReport}{Questionnaire}{SurveyAnswer}{3,537}
  \ivCard[at={(Player |- Word)}]{SItem}{ivSelfReport}{Questionnaire}{SurveyItem}{131 (55 / 26 / 50)}

  \coordinate (J) at ($(Player.east)!0.5!(Track.west)$);
  \draw[iv has] (Experiment) -- (Player);
  \draw[iv has] (Player) -- (Track);
  \draw[iv has] (J) |- (Audio.west);
  \draw[iv has] (J) |- (Word.west);
  \fill[ivEdge] (J) circle (0.9pt);
  \draw[iv transcribed] (Audio) -- (Word);
  \draw[iv has] (Player) -- (QW);
  \draw[iv has] (IItem) -- (QW);
  \draw[iv has] (Experiment.west) -- ++(-2.2mm,0) |- (SAns.west);
  \draw[iv has] (SItem) -- (SAns);

  \ivBadge{Experiment}{1} \ivBadge{Player}{2} \ivBadge{Track}{2} \ivBadge{Audio}{2}
  \ivBadge{QW}{2} \ivBadge{SAns}{2\,+\,3} \ivBadge{SItem}{2\,+\,3}

  \coordinate (Margin) at ([xshift=-2.6mm]SAns.west);
  \begin{scope}[on background layer]
    \node[iv group, fill=black!6, inner sep=1.6mm, fit=(Experiment)(Track)(SAns)(Word)(Margin)] (VR) {};
  \end{scope}
  \node[iv grouplabel] at ([yshift=0.3mm]VR.north west) {VR INTERACTION DATA};

  \ivSetCardWidth{2.45cm}
  \ivCard[right=10mm of Track, yshift=-1.3mm]{Ref}{ivLanguage}{Post-edited}{ReferenceWord}{54 recordings,\\[-0.4ex] 57,347 tokens}
  \ivCard[at={(Ref |- Word)},yshift=1mm]{Neg}{ivAnnotation}{Cue \& scope}{Negation}{1,422 sentences,\\[-0.4ex] 342 instances}
  \ivBadge{Ref}{4} \ivBadge{Neg}{5}
  \begin{scope}[on background layer]
    \node[iv group, fill=black!6, inner sep=1.6mm, fit=(Ref)] (RefG) {};
    \node[iv group, fill=black!6, inner sep=1.6mm, fit=(Neg)] (NegG) {};
  \end{scope}
  \node[iv grouplabel] at ([yshift=0.3mm]RefG.north west) {REFERENCE TRANSCRIPTS};
  \node[iv grouplabel] at ([yshift=0.3mm]NegG.north west) {NEGATION ANNOTATIONS};

  \coordinate (X) at ($(VR.east)!0.5!(RefG.west)$);
  \draw[iv flow] (Word.east) -- (Word.east -| X) |- (Ref.west);
  \draw[iv flow] (Word.east -| X) -- (NegG.west |- Word.east);
  \draw[ivEdge, line width=0.5pt] (Audio.east) -- (Audio.east -| X);
  \fill[ivInk] (Audio.east -| X) circle (1.0pt);
  \fill[ivInk] (Word.east -| X) circle (1.0pt);

  \coordinate (VRtop) at ([yshift=2.6mm]VR.north);
  \begin{pgfonlayer}{ivback}
    \node[iv group, inner sep=1.6mm, fit=(VR)(RefG)(NegG)(VRtop)] (Corpus) {};
  \end{pgfonlayer}
  \node[iv grouplabel, font=\fontsize{6.2}{7}\selectfont\bfseries, text=black!75] at ([yshift=0.6mm]Corpus.north west)
    {INTERVIEW-C \textperiodcentered{} released corpus, 27 interviews on one shared session clock};

  \node[iv step, text width=2.05cm, anchor=north east] (S1) at ([xshift=-4.2mm]Corpus.north west)
    {\hspace*{1.8mm}\textbf{Setup}\\ briefing; avatar choice};
  \node[iv step, text width=2.05cm, below=2.2mm of S1] (S2)
    {\hspace*{1.8mm}\textbf{VR interview}\\ Meta Quest Pro; items read out, answers entered; median 21.7\,min};
  \node[iv step, text width=2.05cm, below=2.2mm of S2] (S3)
    {\hspace*{1.8mm}\textbf{Post-interview}\\ experience questionnaire; interviewer protocol};
  \draw[iv flow] (S1) -- (S2);
  \draw[iv flow] (S2) -- (S3);
  \draw[iv flow] (S2.east) -- (S2.east -| Corpus.west);
  \node[iv grouplabel] at ([yshift=0.6mm]S1.north west) {STUDY};
  \foreach \n/\s in {1/S1, 2/S2, 3/S3}
    \node[iv num] at ([xshift=1.2mm,yshift=-1.2mm]\s.north west) {\n};
\end{tikzpicture}
\caption{Study flow (left: the three stages of each session) and the InterView-C corpus (right), whose three released parts share one session clock. \emph{VR interaction data}: records per table (fingers tracked in 515,995 / 544,280 frames; SurveyItem: 55 interview, 26 protocol and 50 experience items); thin arrows are record links, the dashed arrow marks automatic transcription and thick arrows mark derivation. The \emph{reference transcripts} are post-edited from the automatic transcript against the withheld audio and the \emph{negation annotations} are sampled from it. Badges give the creating stage; 4 (post-editing) and 5 (annotation) continue the study stages.}
\label{fig:flow}
\end{figure*}
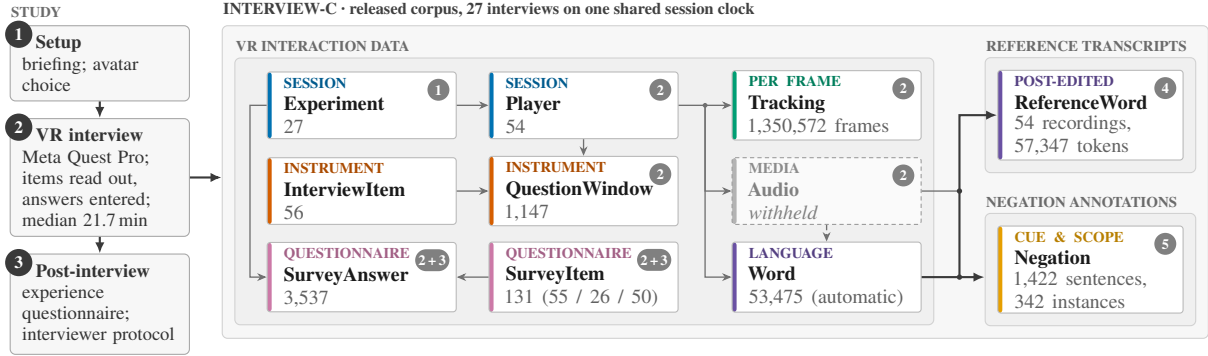

With InterView, we conducted 27 survey interviews in which interviewer and interviewee meet as avatars in a shared virtual room, while the headsets of both record their gaze, facial expressions and head, body, hand and finger movements alongside their speech.
Recording these signals addresses requirement (a), but not (b): only a few current models process nonverbal channels, and even models described as multimodal typically accept text, images and audio~\citep{Yin:et:al:2024,Abouelenin:et:al:2025}, but not gaze, facial expressions or body motion.
Spoken interaction is therefore still analyzed by passing ASR output to text-based NLP methods, which can fail at two points: ASR can distort linguistically relevant information~\citep{Maheshwari:et:al:2025,Koenecke:et:al:2024} and models trained on existing text resources may not transfer reliably to transcribed speech, as observed for negation detection~\citep{Hammerla:et:al:2025b}.
Human-verified transcripts and linguistic annotations that share the time base of the tracking data provide a reliable interface between the two and keep text-based analyses aligned with the nonverbal behavior.
Our contribution is the InterView-C corpus, which consists of three parts (Figure~\ref{fig:flow}):
\begin{enumerate}
    \item \textbf{VR interaction data} (Section~\ref{sec:dataset}): for all 27 interviews, synchronized tracking of both speakers (eyes, head, body, face, hands and fingers), word-timed automatic transcripts, time windows of the questionnaire items and the questionnaire responses.
    \item \textbf{Reference transcripts} (Section~\ref{sec:transcription}): a manually post-edited verbatim transcript of all 54 recordings (57,347 tokens), which we use to evaluate nine open-weight ASR systems and to analyze their errors on survey interview speech.
    \item \textbf{Negation annotations} (Section~\ref{sec:neg_main}): negation cue and scope labels for 1,422 transcribed sentences, with a comparison between training on InterView-C and cross-corpus transfer from existing negation resources.
\end{enumerate}
\section{Related Work} \label{sec:related_work}
\paragraph{Interview Mode and Virtual Interviewers.}
Technology-mediated interviews range from live video between human interlocutors to virtual and LLM-based interviewers, with prior work examining disclosure, data quality and respondent experience across these settings~\citep{Endres:et:al:2023,Conrad:et:al:2023,Lind:et:al:2013,Lucas:et:al:2014,Barari:et:al:2026,Lang:et:al:2025,Leybzon:et:al:2025,Krajcovic:et:al:2026}. 
These studies show that both interview modality and the perceived nature of the interviewer can affect disclosure and response behavior. 
More recent systems increasingly support conversational and adaptive interviewing, including LLM-based chatbots, voice agents and embodied virtual interviewers. 
Interviews between avatars have also been studied in desktop virtual worlds and qualitative VR settings~\citep{Hasler:et:al:2013,Young:et:al:2024}, but, to our knowledge, no survey corpus combines two human interlocutors represented as avatars with headset tracking of both parties.
\paragraph{Multimodal Interaction Corpora.}
Multimodal corpora of face-to-face interaction range from meetings~\citep{McCowan:et:al:2005} and acted and improvised emotional dialogue~\citep{Busso:et:al:2008} to German route descriptions annotated for speech-gesture alignment~\citep{Luecking:et:al:2013}.
Motion-capture corpora record body and finger motion together with each speaker's audio in dyadic conversations~\citep{Lee:et:al:2019} and recent corpora have expanded to include hundreds or thousands of hours~\citep{Agrawal:et:al:2025,McLean:et:al:2025}.
These corpora are recorded in physical rooms using cameras or marker-based capture.
Data recorded with VR headsets, which capture gaze and facial expressions in the context of avatar animation, are still rare~\citep{Rack:et:al:2023}.
For instance, \citet{Kim:Lee:2025} use the data to predict social connection based on turn-taking, gaze and nodding in dyadic VR conversations.
Furthermore, none of these corpora contain structured survey interviews or pair the interaction with the interviewee's survey answers.
Releasing such data requires care, since head and hand motion alone suffices to re-identify VR users~\citep{Nair:et:al:2023}.
\paragraph{Speech Recognition for Conversational and Survey Speech.}
ASR systems that perform well on read-speech benchmarks have been shown to degrade substantially on conversational speech, with errors correlating with disfluencies~\citep{Maheshwari:et:al:2025}, with Whisper-based systems hallucinating content, particularly over non-speech audio~\citep{Koenecke:et:al:2024}.
Although ASR is already used to transcribe and monitor survey interviews~\citep{Sun:Yan:2023,WollinGiering:etal:2024}, its accuracy has been assessed on long, open-ended speech, not on the short closed answers of interviewer--interviewee dialogue on which standardized survey data rest.
\paragraph{Negation in Spoken and Multimodal Interaction.} 
Negation has received comparatively limited attention in computational linguistics and remains underrepresented in widely used NLU resources~\citep{JimenezZafra:et:al:2020,Hossain:et:al:2022}. 
Existing work has primarily focused on cue and scope detection in written text, with dedicated datasets concentrated on English and comparatively few resources available for other languages, including German~\citep{JimenezZafra:et:al:2020,Hartmann:et:al:2021}. 
This gap is particularly pronounced for spoken and conversational language, for which annotated negation corpora are scarce.
More recent work has considered negation in spoken and multimodal interaction, including VR settings~\citep{AbuSaleh:et:al:2026,Hammerla:et:al:2026}. 
Moreover, previous cross-corpus evaluations have found substantial variation in negation-detection performance across datasets~\citep{Hammerla:et:al:2025b}, further motivating the development of annotated resources for German spoken and interactional language.
\section{Experiment Setup} \label{sec:experiment}
\begin{figure}
    \centering
    \includegraphics[width=\linewidth]{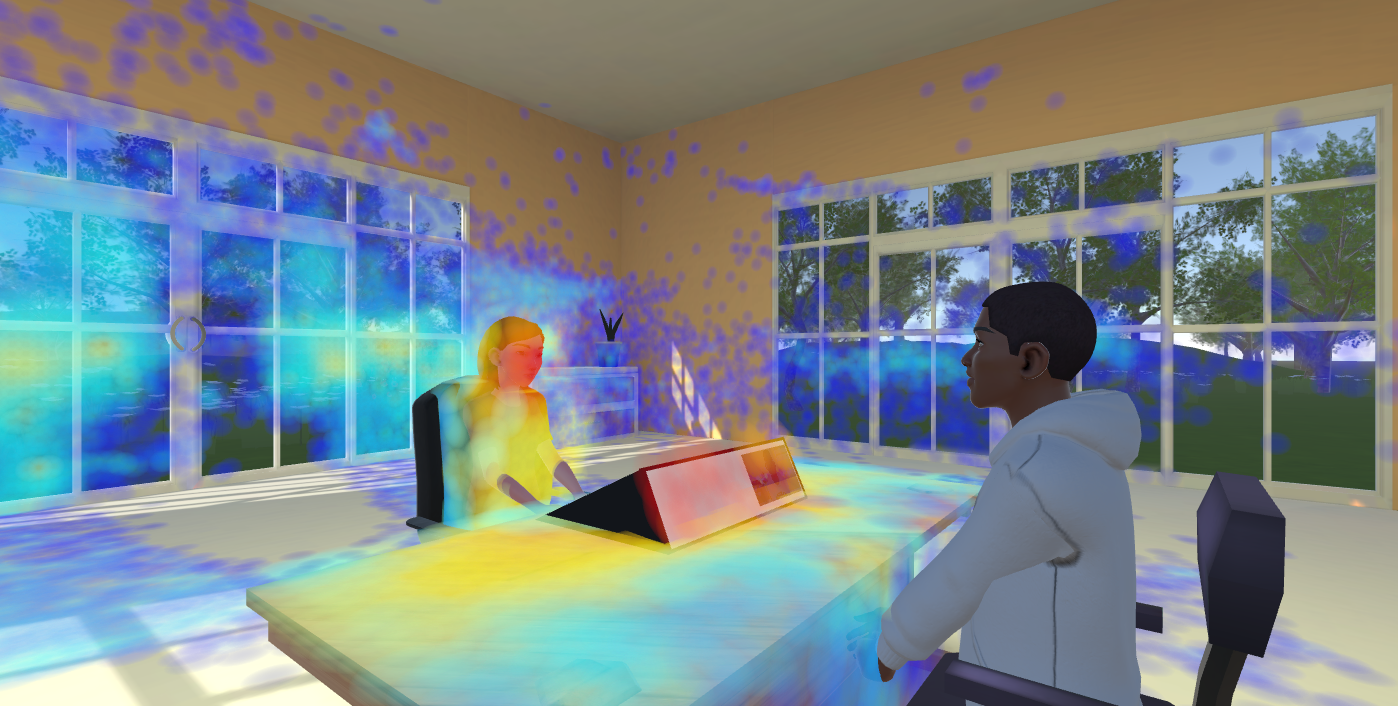}
    \caption{
    The virtual interview environment with the interviewee's 3D gaze points, colored by dwell time from blue (shorter) to red (longer).
    }
    \label{fig:room}
\end{figure}

\paragraph{Procedure.}
Each interviewee was first briefed by an instructor.
The participant then chose one of four avatars (A-D) to represent them in VR~(Appendix~\ref{app:avatars}).
One of the remaining three avatars was then assigned to the interviewer.
The interviewee then put on the Meta Quest Pro VR headset, after which the instructor left the room and the interview began.
The interviewer and interviewee were in different locations and only met in VR.
The interviewees had not seen their interviewer in person beforehand.
The three interviewers were members of the project team and had no prior experience conducting interviews.
\paragraph{Virtual room.}
The interview was conducted in a virtual room that contained three large windows as well as some cupboards, plants and a couch (Figure~\ref{fig:room}).
At the center of the room stood a large office desk, with the interviewee seated at one end and the interviewer at the other.
A browser panel on the desk, facing the interviewer, displayed the LimeSurvey questionnaire, which the interviewer operated during the interview.
The side facing the interviewee showed a mirror in which they could see their own avatar, imitating the self-view of a video interview and strengthening the sense of embodiment~\citep{GonzalezFranco:et:al:2010}.
For items with response scales, the answer options were also displayed to the interviewee on the same panel.

\paragraph{Recording.}
During the interview, all data the Meta Quest Pro can capture were recorded, for both the interviewer and interviewee. 
This includes gaze, face, body, hand and finger tracking alongside the audio.
The main questionnaire is listed in Appendix~\ref{app:survey_items}; all 131 items are part of the release.
\paragraph{Questionnaire.}
The interview was conducted in German.
The interviewer read out a standardized questionnaire of 52 items: 37 asked in every interview, 11 only when a previous answer made them applicable and four income follow-ups that were never triggered.
Most of the items were taken from the German National Education Panel Study (NEPS)~\citep{Blossfeld:Rossbach:2019}, listed in Appendix\ref{app:survey_items}.
The questionnaire covers life satisfaction, partnership and family, education and employment, earned income, locus of control and social belonging and trust.
Since we wanted to learn how the avatar setting affects answers that may cause discomfort in conventional interviews, the questionnaire deliberately includes items known to be sensitive, such as income and relationship conflicts~\citep{Tourangeau:Yan:2007,Yan:2021} and scales on socially aversive traits such as narcissism and Machiavellianism~\citep{Paulhus:Williams:2002}.
The interview was designed to take about 20 minutes in total to complete.
After the interview, the interviewee completed an experience questionnaire on the VR setting, the avatars and honesty (35 items for every respondent plus up to 7 open follow-ups) and the interviewer completed a protocol of the interview (11 items plus up to 14 follow-ups on technical problems and the respondent's discomfort).

\section{Corpus} \label{sec:dataset}
InterView-C contains 27 German-language survey interviews, each conducted entirely in VR, with both the human interviewer and human interviewee represented by avatars.
For every interview, the corpus comprises three parts (Figure~\ref{fig:flow}, Appendix Table~\ref{tab:schema}):
(1)~the \emph{VR interaction data}, i.e., synchronized tracking streams of both interlocutors (gaze, head, body, face, hands and fingers), word-timed automatic transcripts of both speakers, time windows for each questionnaire item and the questionnaire data;
(2)~the \emph{reference transcripts}, a manual verbatim post-edit of the automatic transcripts (Section~\ref{sec:transcription}); and
(3)~the \emph{negation annotations}, cue and scope labels on a sample of the transcribed sentences (Section~\ref{sec:neg_main}).
All layers are timestamped, enabling, for example, the interviewee's gaze and facial expressions during a spoken response to be retrieved in a single query.
This section gives the overall composition and statistics and describes the VR interaction data; the two human-annotated parts are described in Section~\ref{sec:tasks}.
The raw audio is not part of the corpus (Section~\ref{sec:identification}).

\subsection{Composition and Scope}
Each interview is composed of two speakers: one of the 27 interviewees and one of the three interviewers, who conducted 19, 6 and 2 interviews, respectively.
According to the interviewer protocol, 19 interviewees were male and 8 female.
The participants received no compensation and were all highly educated with ages ranging from 18 to 60.
The interviewees chose avatars A, B and C 13, 11 and 3 times, respectively, while avatar D was never chosen.
The avatar representing the interviewer was chosen from among the remaining three avatars, once the interviewee had selected their own avatar.
This process was not randomized, resulting in six interviews with avatar A, five interviews with avatar B, eight interviews with avatar C and eight interviews with avatar D.

\subsection{Corpus Statistics}
Interviews last a median of 21.7 minutes. 
The questionnaire itself takes a median of 13.1 minutes when measured from the first to the last administered item. 
The remaining time is spent on setup, introduction, instructions and closing statements.
Since interviewers read every item and its answer options, they produce most of the speech.
In the automatic transcript, they produce a median of 1,315 words per interview (range 1,182--2,083), against 420 (115-1,564) for the interviewees.
The high variance on the interviewee side reflects the mix of closed items, which typically receive one- to three-word answers and open-ended items, which elicit substantially longer responses depending on the participant.
Of the 3,537 questionnaire answers, 2,440 correspond to items that were actually asked, of which 2,346 received a substantive answer, 64 no answer and 30 with the label \enquote{don't know}; the remaining 1,097 are items that were not asked (filter skips, break-offs or missing questionnaires).
Table~\ref{tab:statistics} gives an overview.
\begin{table*}[t]
    \small
    \centering
    \begin{tabularx}{\textwidth}{Xrrr}
         & Interviewee & Interviewer & Total \\
        \midrule
        Speakers                                  & 27                 & 3                    & 30 \\
        Tracked time (h)                          & 9.9                & 10.5                 & 20.4 \\
        Tracking frames                           & 654,682            & 695,890              & 1,350,572 \\
        Word tokens, automatic (\texttt{Word}) & 14,772             & 38,703               & 53,475 \\
        Word tokens, post-edited reference (\texttt{ReferenceWord})$^\ddagger$ & 15,573 & 40,191 & 55,764 \\
        Words per interview, automatic, median (range) & 420 (115--1,564)   & 1,315 (1,182--2,083) & -- \\
        Item windows (located in transcript)      & --                 & --                   & 1,147 \\
        Questionnaire answers (substantive / total)$^\dagger$ & --  & --                   & 2,346 / 3,537 \\
        \bottomrule
    \end{tabularx}
    $^\ast$Scoring tokenization of Section~\ref{sec:transcription}, including 1,432 filled pauses. 
    $^\dagger$131 items per interview: 55 interview items and 50 experience items answered by the interviewee and 26 protocol items answered by the interviewer.
    $^\ddagger$Released tokens as written by the annotators, without \texttt{[unknown]} markers; 55,596 (99.7\%) are forced-aligned.
    \caption{General statistics about the data InterView-C contains.}
    \label{tab:statistics}
\end{table*}

\subsection{Synchronization and Tracking Data}
All streams of both headsets are placed on one session clock: each headset’s clock is mapped onto a shared clock with a per-session offset and sub-second timing follows the audio sample clock (Appendix \ref{app:sync}).
Tracking is logged at a median of 18.4Hz, which suffices for gaze, head, hand and facial behavior at the level of words and turns but not for saccades or brief facial events.
The hands were tracked in 38.2\%(left) and 40.3\%(right) of the frames; untracked frames are flagged.
\subsection{De-identification and Release} \label{sec:identification}
All tracking streams, both transcript layers, the item windows, the questionnaire data and both task resources are available on request at huggingface\footnote{\url{https://hf.co/datasets/AnonymousKiwi/InterView-C}} under a data use agreement that restricts use to non-commercial research and prohibits re-identification; the raw audio is withheld.
The code is released under AGPLv3 on GitHub\footnote{\url{https://github.com/TheBv/faces-dataset-pre-study}}.
Transcripts are verbatim with word-level timestamps; names of persons, places and institutions are replaced by typed placeholders.
Head and hand motion alone can re-identify VR users~\citep{Nair:et:al:2023,Miller:et:al:2020} and the release adds gaze, facial blendshapes and answers on gender, partnership, education, occupation and income.
Re-identification from motion, however, requires reference recordings of the same person and the corpus contains no names, voices or images; with restricted access, this lowers but does not eliminate the risk.
Participants gave informed consent to the recording and to the release of the de-identified data for research purposes.

\section{Human Annotation and Evaluation}\label{sec:tasks}
The two human-annotated parts of InterView-C target the speech and text channel, through which most current models access spoken interaction.
For each part, we describe its creation and evaluate current models on it: the reference transcripts serve as an ASR benchmark (Section~\ref{sec:transcription}) and the negation annotations as training and test data for negation cue and scope detection (Section~\ref{sec:neg_main}).
\subsection{Reference Transcripts and ASR Evaluation} \label{sec:transcription}
Survey interviews are a demanding setting for ASR:
Interviewers largely read out scripted items and answer options, resembling the read speech that dominates ASR training data, but embedded in a dialogue: each headset recording contains long silences while the other person speaks, interspersed with short backchannels (\textit{mhm}, \textit{ja}) and filled pauses.
Interviewees, in contrast, speak spontaneously, often answering with a single word, and their answers contain disfluencies such as filled pauses, repetitions and self-repairs, possibly reflecting uncertainty in the unfamiliar setting.
We therefore manually corrected the transcripts of all 54 recordings (one per speaker and interviewer) and use them to benchmark current open-weight ASR models.
\paragraph{Reference transcripts.}
We first transcribed silence-delimited segments with CrisperWhisper~\citep{Zusag:et:al:2024}, a Whisper model fine-tuned for verbatim transcription, using beam search.
It was chosen for its strong German results and word-level timestamps.
Five annotators post-edited them in a web interface with audio playback, one annotator per recording.
The transcripts are verbatim: filled pauses (\textit{äh}, \textit{ähm}, \textit{mhm}), repetitions and false starts are kept and unintelligible words are marked as \texttt{[unknown]}.
Numbers up to thirteen and colloquial elisions are written out.
The final reference contains 57,347 tokens, including 1,432 filled pauses (2.5\%) and 380 immediate word repetitions.
About four in five filled pauses were added by annotators, and post-editing changed around 9.1\% of the words of the automatic transcripts.
Forced alignment with a German wav2vec~2.0 model~\citep{Baevski:et:al:2020,Wang:et:al:2021} assigns timestamps to 99.7\% of the reference words.
Mapped onto the session clock, these words form the ReferenceWord layer of the corpus.
\paragraph{Experimental setup.}
We evaluate nine systems zero-shot under three decoding regimes (Table~\ref{tab:asr}).
Three systems decode the full recordings with their own long-form strategy: Whisper large-v3 and large-v3-turbo~\citep{Radford:et:al:2023} and CrisperWhisper~2~\citep{Wagner:et:al:2026} in verbatim mode.
WhisperX~\citep{Bain:et:al:2023} uses its built-in Voice Activity Detection (VAD) and runs Whisper large-v3 on the resulting segments.
The other five receive voice-activity segments of at most 30\,s from Silero VAD~\citep{Silero:2024}: Qwen3-ASR-1.7B~\citep{Shi:et:al:2026}, Voxtral Mini 3B~\citep{Liu:et:al:2025}, Parakeet-TDT-0.6B-v3 and Canary-1B-v2~\citep{Sekoyan:et:al:2025} and Phi-4-multimodal~\citep{Abouelenin:et:al:2025}.
To vary the decoding strategy, we also ran CrisperWhisper~2 on the VAD segments in verbatim and intended (clean) modes.
We further ran Whisper large-v3-turbo on the full recordings without conditioning on previously decoded text.
Where a model exposes a language option, it is set to German.
Before scoring, we normalize both texts: we lower-case them, remove punctuation, spell out numbers, expand abbreviations and elisions, map filled-pause tags to filled pauses, drop other non-speech tags (\textit{[laughter]}) and treat open, hyphenated and closed compounds alike (\textit{Zoom Call}, \textit{Zoom-Call}, \textit{Zoomcall}).
Our main measure, word error rate (WER), also removes filled pauses and cut-off word fragments (\textit{hau-}) from both sides, since only verbatim systems produce them.
The verbatim WER keeps the filled pauses.
We report WER pooled over all recordings, with 95\% confidence intervals from a bootstrap over interviews (1,000 resamples).
Because WER treats all words alike and agrees poorly with human judgments of transcript quality, we also report character error rate (CER) and semantic distance (SemDist; \citealp{Kim:et:al:2021}), which agrees best with human judgments in comparative studies~\citep{Kim:et:al:2022}.
SemDist is one minus the cosine similarity between multilingual E5 embeddings~\citep{Wang:et:al:2024} of each reference line and the hypothesis words aligned to it, averaged over lines.

\paragraph{Results and analysis.}
Excluding the automatic transcript the reference was derived from, WhisperX achieves the lowest WER (8.0\%), followed by Qwen3-ASR (10.4\%).
Qwen3-ASR is in turn better than Canary (1.1 points, CI 0.8--1.4) and slightly better than Voxtral (0.4, CI 0.0--0.8), whereas Voxtral, Parakeet, Phi-4 and Canary cannot be separated: all differences between adjacent systems in this group have CIs that include zero.
\begin{table*}[t]
    \centering
    \small
    \setlength{\tabcolsep}{4.5pt}
    \begin{tabular}{lrrrrrrrrrrr}
    System & WER [95\% CI] & CER & SemD & IR & IE & Verb. & Del. & Ins. & Cue & Num & Hall./h \\
    \midrule
    CrisperWhisper$^\dagger$ & 7.4 [6.3--8.6] & 5.8 & 1.72 & 6.2 & 10.5 & 9.1 & 4.0 & 1.6 & 94.2 & 86.3 & 2.2 \\
    \midrule
    \multicolumn{12}{l}{\emph{Own VAD, batched decoding}} \\
    WhisperX (large-v3) & \textbf{8.0} [7.4--8.7] & \textbf{5.5} & \textbf{1.60} & 6.6 & 12.0 & 10.0 & 4.0 & 1.4 & 93.2 & 86.8 & 1.0 \\
    \midrule
    \multicolumn{12}{l}{\emph{Shared Silero VAD segments}} \\
    Qwen3-ASR 1.7B & 10.4 [9.4--11.6] & 6.1 & 1.90 & 8.6 & 15.4 & 12.4 & 3.3 & 2.0 & 90.8 & 84.5 & 1.1 \\
    Voxtral Mini 3B & 10.8 [9.9--11.8] & 6.7 & 2.16 & 8.5 & 17.3 & 12.6 & 3.5 & 2.1 & 90.4 & 77.6 & 1.3 \\
    Parakeet-TDT 0.6B v3 & 11.1 [10.0--12.1] & 6.3 & 2.32 & 8.6 & 17.9 & 12.8 & 2.9 & 2.3 & 90.4 & 72.7 & 1.4 \\
    Phi-4-multimodal & 11.2 [9.9--12.5] & 6.3 & 1.93 & 9.2 & 16.5 & 13.5 & 2.8 & 2.5 & 88.1 & 80.6 & 1.4 \\
    Canary 1B v2 & 11.5 [10.4--12.6] & 6.9 & 2.21 & 8.9 & 18.6 & 13.3 & 3.6 & 2.2 & 87.0 & 82.1 & 1.6 \\
    CrisperWhisper 2 (verbatim) & 12.8 [11.3--14.3] & 7.4 & 2.48 & 10.6 & 18.8 & 15.5 & 3.9 & 2.5 & 88.8 & 75.5 & 1.2 \\
    \midrule
    \multicolumn{12}{l}{\emph{Full recordings (long-form decoding)}} \\
    Whisper large-v3-turbo$^\ddagger$ & 12.8 [11.5--14.2] & 8.9 & 1.99 & 10.6 & 18.9 & 14.7 & 5.0 & 3.3 & 88.6 & 86.4 & 3.6 \\
    CrisperWhisper 2 (verbatim) & 22.2 [17.8--27.5] & 14.2 & 2.52 & 18.5 & 32.2 & 24.4 & 6.0 & 9.7 & 86.0 & 74.0 & 3.4 \\
    Whisper large-v3 & 44.5 [40.7--48.9] & 37.8 & 5.55 & 41.3 & 53.2 & 45.3 & 13.0 & 16.7 & 76.1 & 61.9 & 24.5 \\
    \end{tabular}
    \caption{ASR results on all 54 recordings (\%, SemDist $\times 100$; lower is better except for recall). \emph{WER}: WER without filled pauses; \emph{CER}: character error rate; \emph{SemD}: mean semantic distance per reference line; \emph{IR}/\emph{IE}: WER on interviewer/interviewee recordings; \emph{Verb.}: WER with filled pauses kept; \emph{Del.}/\emph{Ins.}: deletions/insertions relative to the reference length; \emph{Cue}/\emph{Num}: recall of negation cues and number words (Section~\ref{sec:transcription}, Results); \emph{Hall./h}: inserted passages of at least five words per hour of audio. $^\dagger$The reference was post-edited from this transcript and is shown for reference only. $^\ddagger$Without conditioning on previously decoded text.}
    \label{tab:asr}
\end{table*}
CER and SemDist largely reproduce this ranking but rate WhisperX better than the automatic transcript, which WER favors because the annotators kept its word forms.
For the VAD-based systems, WER on the spontaneous and often very short interviewee answers is 1.8 to 2.1 times that on the largely scripted interviewer speech.
The decoding strategy matters more than the model.
Whisper's default long-form decoding conditions each window on previously decoded text, which carries repetition loops across the silences~\citep{Koenecke:et:al:2024}: Whisper large-v3 reaches 44.5\% WER on the full recordings (24.5 inserted passages per hour), compared with 8.0\% for the same model on VAD segments, and large-v3-turbo 30.1\%.
Without conditioning, large-v3-turbo reaches 12.8\%, and CrisperWhisper~2 improves from 22.2\% to 12.8\% on VAD segments.
In verbatim mode, CrisperWhisper~2 is the only system that transcribes most filled pauses (64--78\% recall vs.\ at most 30\%); its intended mode misses more short answers (13.8\% WER).
Errors concentrate where survey research is most sensitive.
Short answers are lost: every system misses 13--23\% of single-word reference lines (CrisperWhisper~2 on VAD segments up to 30\%), mostly scale points or \textit{ja}/\textit{nein}, against at most 1\% of lines with ten or more words for the VAD-based systems, and number-word recall ranges from 62\% to 87\%.
On short segments, several models also switch language or script, and study-specific vocabulary is often garbled.
Since the reference was post-edited from CrisperWhisper output, it favors that transcript: 418 reference tokens (0.73\%) are reproduced only by it while most other systems agree on an alternative, so we do not rank it against WhisperX (0.6 points WER apart).
The annotators also differ in thoroughness (Appendix~\ref{app:asr}) and as each recording was corrected by a single annotator, inter-transcriber agreement cannot be reported.
\subsection{Negation Annotation and Detection} 
\label{sec:neg_main}
The third part of InterView-C annotates negation, a fine-grained linguistic phenomenon on which survey answers often hinge.
Negation resources for conversational language are scarce and negation detection shows substantial cross-corpus variation~\citep{Hammerla:et:al:2025b}, so we annotate negation in the transcribed dialogue, train models on these annotations and compare them with existing models.
As a second downstream use case, we investigate whether InterView-C supports the analysis of a fine-grained linguistic phenomenon, using negation as a case study. 
We train negation models on our annotations and assess how released models trained on existing negation corpora perform in this new annotation and data setting.
Given the limited availability of negation resources for conversational language and the poor cross-domain generalization observed in previous work~\citep{Hammerla:et:al:2025b}, our corpus provides an opportunity to study negation in transcribed dialogue. 
Following \citet{Morante:Blanco:2012} and \citet{JimenezZafra:et:al:2020}, the task consists of two parts: \emph{cue detection} identifies words that express negation (e.g., \textit{nicht}, \textit{kein}, \textit{nie}), while \emph{scope detection} identifies, for a given cue, the words that fall within its semantic scope.
\paragraph{Sample and annotation.}
We sampled 1,422 sentences from all 27 interviews (52--53 per interview) without conditioning the sampling procedure on negation cues or labels.
The sample is stratified by speaker role and conversational position and contains at most one sentence from each cluster of exact or near-duplicate sentences, reducing the influence of repeated scripted questions.
Splits are defined over complete interviews (21/3/3 for train/validation/test), so evaluation measures generalisation to unseen conversations and interviewees (Appendix~\ref{app:neg-sampling}).
Three annotators marked negation cues and linked each cue to its scope, with every sentence independently annotated by two annotators (Appendix~\ref{app:neg-guidelines}).
Cue agreement is high (Krippendorff's $\alpha=0.87$), while scope agreement on instances with identical cues is somewhat lower ($\alpha=0.81$), with disagreements primarily concerning scope extent (Appendix~\ref{app:neg-agreement}).
\paragraph{Reference annotation.}
We consolidate the two annotations automatically: cue annotations are unioned to favor recall, agreed scopes are retained and disputed scope tokens are resolved with a Dawid--Skene model~\citep{Dawid:Skene:1979} estimated on the training split only; stricter cue annotations and union- and intersection-based scope alternatives are released as well (Appendix~\ref{app:neg-consolidation}).
The reference contains 342 negation instances in 281 sentences (19.8\%; Table~\ref{tab:neg-ds}).
The cues are dominated by \textit{nicht} (54\%), followed by \textit{nie}, response particles (\textit{nein}, \textit{nee}) and forms of \textit{kein}; 58 cue types occur in total.
\begin{table}[t]
\centering
\small
\begin{tabular}{lrrr}
& Train & Val & Test \\
\midrule
Interviews            & 21    & 3     & 3     \\
Sentences             & 1,106 & 158   & 158   \\
Tokens                & 9,464 & 1,531 & 1,750 \\
Sentences w/ negation & 225   & 29    & 27    \\
Negation instances    & 267   & 38    & 37    \\
Cue tokens            & 278   & 40    & 38    \\
Scope tokens$^\ddagger$ & 969 & 123   & 190   \\
\bottomrule
\end{tabular}
$^\ddagger$Excluding cue tokens.
\caption{Consolidated negation annotation per split.}
\label{tab:neg-ds}
\end{table}
\paragraph{Experimental Setup.}
We use the syntax-aware \texttt{D-Neg} model of \citet{Hammerla:et:al:2025}, based on \texttt{EuroBERT-210m}~\citep{Boizard:et:al:2025}. 
It augments transformer representations with part-of-speech and dependency-relation embeddings and applies a graph attention layer over dependency trees predicted with \texttt{spaCy}'s \texttt{de\_core\_news\_sm} model~\citep{Honnibal:et:al:2023}. 
For scope detection, cue words are replaced by \texttt{[CUE]}.
We train \texttt{D-Neg} on our training split for 10 epochs using the original hyperparameters (learning rate $2\cdot10^{-5}$, batch size 16, weight decay 0.01), select the best validation-F1 checkpoint, and repeat training with three seeds (42--44).
For comparison, we evaluate all released German GAT variants of \texttt{D-Neg}, trained on German versions of \texttt{ConanDoyle-neg}~\citep{Morante:Daelemans:2012}, \texttt{SFU Review}~\citep{Konstantinova:2012}, \texttt{BioScope}~\citep{Vincze:2008}, \texttt{SOCC}~\citep{Kolhatkar:2019}, and \texttt{DT-Neg}~\citep{Banjade:Rus:2016}, without further fine-tuning.
Because the source corpora also differ in annotation conventions and were translated into German, while our model is trained using the same annotation and consolidation procedure as the test reference, this comparison does not isolate spoken--written domain transfer; it reflects these factors jointly.
All models are evaluated on the same held-out test split; scope detection uses gold cues. We report word-level precision, recall, and F1 for the positive class, plus exact-match accuracy for complete cue and scope predictions. 
For our trained model, we report mean and standard deviation across seeds.
\paragraph{Results and analysis.}
Our in-domain \texttt{D-Neg} model outperforms all released GAT variants on both tasks (Table~\ref{tab:neg-results-main}), reaching 0.88 F1 for cue detection and 0.66 for scope detection, compared with 0.82 and 0.61 for the strongest transferred models.
Transfer performance varies strongly across source corpora: cue models trained on user-generated text (\texttt{SFU Review}, \texttt{SOCC}) perform well, as does the \texttt{ConanDoyle-neg} model, whose annotation guidelines we follow, whereas \texttt{BioScope} reaches at most 0.58 F1. For scope detection, \texttt{DT-Neg} transfers best.
The \texttt{DT-Neg} cue model is an outlier (0.11 F1), frequently misclassifying words such as \textit{ich} and \textit{ja}.
Most cue differences lie in the long tail: all models detect nearly all instances of \textit{nicht} and \textit{nie} (68\% of test cues), whereas our model recovers about eight of the remaining 12 cues, versus two for the strongest transferred model.
The strongest transferred cue models show markedly lower recall than ours (0.71--0.74 vs.\ 0.89), despite higher precision (0.93 vs.\ 0.88).
Scope predictions are conservative: precision exceeds recall (0.81 vs.\ 0.56), predicted scopes are shorter than references (3.6 vs.\ 5.1 tokens), and 32\% are strict subsets versus 14\% strict supersets (see Appendix~\ref{app:neg-detailed-results} for detailed results and error analyses).
\begin{table}[t]
\centering
\small
\begin{tabular}{lcc}
Training data & Cue F1 & Scope F1 \\
\midrule
BioScope (abstracts) & 0.58 & 0.49 \\
BioScope (full) & 0.26 & 0.34 \\
ConanDoyle-neg & 0.81 & 0.53 \\
DT-Neg & 0.11 & 0.61 \\
SFU Review & 0.82 & 0.34 \\
SOCC & 0.79 & 0.38 \\
\midrule
Ours (3 seeds) & \textbf{0.88}{\scriptsize$\pm$0.02}
& \textbf{0.66}{\scriptsize$\pm$0.02} \\
\end{tabular}
\caption{Word-level F1 (\%) on the test split for our and released \texttt{D-Neg} models (mean $\pm$ std. over three seeds). More detailed results are reported in Appendix~\ref{app:neg-detailed-results}.}
\label{tab:neg-results-main}
\end{table}
\section{Conclusion} \label{sec:conclusion}
We presented InterView-C, a corpus of 27 German survey interviews conducted entirely in virtual reality between two human interlocutors represented by avatars.
It combines VR interaction data with synchronized tracking of both speakers and human reference transcripts, as well as negation cue and scope annotations. 
These provide an interface between multimodal interaction and predominantly text-based NLP methods.
Our evaluations show that the points at which such a pipeline can fail are important in practice.
Once decoding is no longer dependent on text decoded before a silence, errors in ASR concentrate on short closed answers, number words, and negation cues -- the tokens on which survey responses depend.
Negation models trained on existing corpora show variable transfer to the transcribed interviews and generally underperform in-domain training, although differences in annotation conventions, translation and label construction mean that this gap cannot be attributed to the spoken-interaction domain alone.
Future work will use the nonverbal streams to investigate how gaze, head movement and facial expression relate to responses on sensitive items and spoken negation.
InterView-C thus enables linguistic analyses of spoken interaction while retaining their alignment with multimodal behavior.
By releasing all three parts, we hope to support research at the intersection of survey methodology, multimodal dialogue and spoken language processing.
\section*{Limitations} \label{sec:limitations}
\paragraph{Sample.}
The corpus is small and unbalanced: 27 interviewees (19 male, 8 female), three interviewers with 19, 6 and 2 interviews, one questionnaire and German only.
The interviewers were members of the project team and the interviewer's avatar was not randomized, so interviewer, avatar and interviewee choice are partly confounded.
Additionally, the interviewees were all highly educated and the interviewers had no prior interviewing experience.
Findings on interview behavior should therefore be treated as exploratory.
\paragraph{Tracking data.}
Tracking is logged at a median of 18.4\,Hz, which is too coarse for saccade-level eye-movement analysis or brief facial events and the hands were tracked in only about 40\% of the frames.
\paragraph{ASR evaluation.}
Because the audio is withheld, third parties cannot evaluate new systems on our reference.
Each recording was post-edited by a single annotator starting from CrisperWhisper output, so the reference favors that system and may retain errors that other systems share.
Additionally, two annotators made noticeably fewer corrections than the others.

\paragraph{Negation subcorpus.}
The test split contains only 37 negation instances, so model differences of a few points are not reliable.
The reference is consolidated automatically (cues by union) and it may contains annotation errors that we identify but do not correct.
\paragraph{Privacy.}
Despite de-identification, motion, gaze and face tracking combined with answers to sensitive questions carry a residual re-identification risk (Section~\ref{sec:identification}).

\section*{Ethics Statement}
All participants gave informed consent for the recording and release of de-identified data for research purposes.
The release withholds the raw audio, replaces directly identifying information in the transcripts with typed placeholders and is shared only on request under a data use agreement that restricts use to non-commercial research and prohibits re-identification.
Logging the tracking at 18.4\,Hz further limits identification from gaze, since eye-movement biometrics degrade markedly at 50\,Hz and below~\citep{Lohr:Komogortsev:2022}.
Head and hand motion, however, remain identifying even when subsampled to 10\,Hz~\citep{Nair:et:al:2024}, so a residual risk remains (Section~\ref{sec:identification}).
The residual re-identification risk of the tracking data is discussed in Section~\ref{sec:identification}.

\section*{Acknowledgments}
We gratefully acknowledge the financial support provided by the German Research Foundation (DFG) for the projects \enquote{New Data Spaces for the Social Sciences} (SPP 2431) - \enquote{FACES} (DFG: 539634240) and \enquote{Negation in Language and Beyond} (CRC 1629) - \enquote{NegLaB} (DFG: 509468465).

\bibliography{custom}

\appendix

\section{Synchronization Details}\label{app:sync}
Each headset stamps all of its streams with its own local clock.
Because some headsets deviated from wall time, each session's clock was mapped onto a shared clock with a constant offset estimated as the minimum of the server's arrival time minus the local time over the session.
Queuing delay is never negative, so the least-delayed packet gives the best estimate.
Using a single constant per session keeps jitter of server arrival times out of the timing.
Sub-second timing comes from the audio sample clock.
The client sends data in flushes, each consisting of an audio chunk and the sensor samples recorded in the same time window.
The client stamps each flush with the local time at the end of the window.
Audio chunks are contiguous between reconnections; therefore, each contiguous run is anchored from the median of the local time minus the cumulative sample time.
This reduces the one-second quantization of the local clock  (standard deviation 0.29\,s per chunk) to about $0.29/\sqrt{n}$\,s for a run of $n$ chunks.
Precision resets only at a reconnect.
Sensor samples are spread over the time window of the audio chunk they were sent with and therefore inherit its precision.
Words are mapped from audio time through the chunks and word durations are taken from the audio sample clock, so tracking frames, words and item windows share one timeline.

\section{Transcription Details}\label{app:asr}
\paragraph{Line placement and alignment.}
Reference lines carry no reliable timestamps since those inserted by annotators have none.
We therefore locate each line through the segment-level systems, whose output is stored per VAD segment.
Every reference word aligned to a hypothesis word by the WER alignment votes for that word's segment. This is pooled over Qwen3-ASR, Parakeet, Canary and Voxtral. 
Each line takes the segments holding at least a quarter of its votes and this is kept monotone over time.
Lines that no system recognizes are placed in an unclaimed segment between their neighbors.
The text of each group of segments is then aligned to its audio using CTC forced alignment and the German wav2vec~2.0 model, which has been fine-tuned on VoxPopuli.
\paragraph{Decoding controls.}
CrisperWhisper~2 in verbatim mode additionally emits word fragments and non-speech tags, which we remove before scoring.
Its long-form output contains eleven inserted runs of 50 or more words, the longest an 886-word loop of \textit{ja das ist ja} in one interviewee recording.
Only 1--5\% of the inserted words of the long-form systems match the other speaker's speech, so the insertions are not explained by cross-talk.
\paragraph{Annotators.}
Table~\ref{tab:asr-annotators} compares the five annotators.
The edit rate is the WER of the automatic transcript against the final reference, i.e., the proportion of words changed by the annotator.
The model WER is the average WER of five VAD-based systems, on the same recordings.
It takes into account both recording difficulty and residual reference errors.
T4 and T5 involved fewer changes and left about twice as many tokens that only the automatic transcript reproduces.
For T5, this coincides with the lowest model WER, suggesting under-correction rather than easier recordings.
\begin{table}[t]
\centering
\small
\setlength{\tabcolsep}{3.5pt}
\begin{tabular}{lcrrrr}
\toprule
 & Rec. (IR/IE) & Edit & Model & Anch. & Low \\
\midrule
T1 & 12 (6/6) & 10.3 & 9.7 & 5.4 & 3.9 \\
T2 & 12 (6/6) & 10.2 & 10.9 & 5.8 & 4.5 \\
T3 & 10 (5/5) & 12.2 & 9.8 & 5.7 & 3.6 \\
T4 & 10 (4/6) & 7.5 & 13.1 & 11.7 & 4.7 \\
T5 & 10 (6/4) & 4.7 & 9.1 & 9.7 & 2.6 \\
\bottomrule
\end{tabular}
\caption{Reference quality per annotator. \emph{Rec.}: recordings (interviewer/interviewee); \emph{Edit}: words changed relative to the automatic transcript (\%); \emph{Model}: WER of the five VAD-based systems (\%); \emph{Anch.}: tokens reproduced only by the automatic transcript, per 1,000 reference tokens; \emph{Low}: lines recognized by at least one system with alignment confidence below 0.05 (\%).}
\label{tab:asr-annotators}
\end{table}
\paragraph{Negation and semantic distance.}
For the 486 distinct reference lines that contain a negation cue and at least four words, we deleted either the first cue or a randomly chosen other word, so both edits cost exactly one WER error.
Deleting the cue increases SemDist ($\times 100$) by 2.35 on average (median 1.54), deleting another word by 1.07 (median 0.54); in 20\% of the lines, however, the cue deletion is rated the smaller change.
Semantic metrics therefore complement, but do not replace, a direct check of negation cues when transcripts feed negation or survey analyses.
\paragraph{Unreferenced speech.}
Of the 55 VAD segments with speech missing from the reference, 32 are in interviewee and 23 in interviewer recordings; most contain short answers (\textit{ja vier}) or remarks about the technical setup.
\section{Negation Cue and Scope Detection}
\label{sec:neg-supp}
\subsection{Negation Annotation}
\label{app:neg-anno}
\subsubsection{Sampling Procedure}
\label{app:neg-sampling}
Dialogue turns from all 27 interviews were automatically segmented into sentences; short interruptions by the other speaker were conservatively bridged, yielding 232 sampled sentences reconstructed across turns.
Candidate sentences were required to contain at least two words, begin with a capital letter and end in terminal punctuation without a truncating ellipsis.
Candidates with unbalanced brackets, transcription artifacts, incomplete syntactic endings or more than 120 words were also removed.
In total, 2,436 candidates failed these form-based checks, leaving 4,220 eligible sentences.
No negation information (cue lexicon, labels or model predictions) was used at any stage of sampling.
We drew a stratified random sample with seed 42.
Each interview contributed 52 or 53 sentences.
Within each interview, the quota was allocated between interviewer and interviewee in proportion to the number of distinct (non-duplicate) sentences available for each role, with a minimum of ten sentences per role and distributed across five chronological positions in the conversation.
To satisfy the duplicate constraint below, positional quotas were relaxed: all five positions are represented, but not equally (448/195/254/241/284 sentences).
All 30 speakers contribute at least ten sentences.
Because many interview questions are scripted, sentences were grouped into duplicate clusters if they were identical after normalization (NFKC, case folding, punctuation removal) or, for sentences of at least six tokens, near-identical (character similarity $\geq 0.92$).
At most one sentence per cluster was sampled across the entire corpus and therefore across splits.
The 4,220 eligible sentences form 2,121 clusters; 232 contain more than one sentence, with the largest containing 46.
Near-duplicate matching is not applied to sentences shorter than six tokens, since a single character can distinguish different answers (\textit{Gar nicht.} vs.\ \textit{Gar nichts.}); 36 near-identical short pairs therefore remain in the sample, nine of them spanning two splits.
Under these constraints, the corpus supports a sample of 1,422 sentences (mean length 9.0 tokens, median 6).
Because duplicate control removes most repeated scripted questions, interviewers contribute 630 sentences (44.3\%) despite producing 69.1\% of the eligible sentences; interviewees contribute 792 (55.7\%).
The three interviewers contribute 83, 210 and 337 sentences; exact balance was infeasible under the joint constraints.
Splits were defined over complete interviews: 21 interviews for training (1,106 sentences), 3 for validation (158) and 3 for testing (158).
Each held-out split contains one interview by each of the three interviewers.
The task therefore evaluates generalization to unseen conversations and interviewees, but not to unseen interviewers.
\subsubsection{Annotation Guidelines}
\label{app:neg-guidelines}
Three annotators labeled the sampled sentences in a web interface.
Sentences were presented one at a time, without surrounding dialogue context, in a seeded random order.
Annotators marked every expression functioning as a negation cue, with cues optionally spanning multiple tokens and linked each cue to the tokens in its semantic scope.
Scopes associated with different cues were allowed to overlap.
Each sentence was assigned to one of the three annotator pairs in rotation, so every sentence received two independent annotations, with each pair assigned exactly one third of the sample (474 sentences).
\subsubsection{Inter-Annotator Agreement}
\label{app:neg-agreement}
Agreement was measured at the token level.
Because no sentence was annotated by all three annotators, we report pairwise Cohen's $\kappa$ together with Krippendorff's nominal $\alpha$, which accommodates missing ratings.
Cue agreement is high ($\alpha=0.87$).
Annotators agree on whether a sentence contains negation at all in 96.6\% of the 1,398 doubly annotated sentences (pooled pairwise $\kappa=0.88$) and their cue annotations are identical for 93.8--96.3\% of shared sentences.
Scope agreement is computed only for negation instances whose cue tokens match exactly between the two annotators (254 instances).
Token-level agreement is $\alpha=0.81$ and complete scopes are identical in 59.3--75.3\% of these instances.
When scopes differ, they differ by three to five tokens on average and the disagreements are largely systematic: A3 marks wider scopes than A1 (94 vs. 15 tokens labeled as scope by only one of the two).
\begin{table}[t]
\centering
\small
\begin{tabular}{lccccc}
& \multicolumn{3}{c}{Cohen's $\kappa$} & & \\
\cmidrule(lr){2-4} \\
Task & A1--A2 & A1--A3 & A2--A3 & $\alpha$ & Exact (\%) \\
\midrule \\
Cue & 0.90 & 0.85 & 0.85 & 0.87 & 93.8--96.3 \\
Scope$^\dagger$ & 0.79 & 0.79 & 0.84 & 0.81 & 59.3--75.3 \\
\bottomrule
\end{tabular}
$^\dagger$Computed only on instances whose cue tokens match exactly (85/86/83 instances per pair).
\caption{
Token-level inter-annotator agreement.
\emph{Exact} gives the range over annotator pairs of the percentage of shared sentences (cue) or matched negation instances (scope) with identical annotations.
}
\label{tab:neg-agreement}
\end{table}
\subsubsection{Consolidation}
\label{app:neg-consolidation}
We derive a single reference annotation automatically rather than through manual adjudication.
\paragraph{Cue consolidation.}
Cue tokens are unioned across annotators, favoring recall; this adds 80 cue tokens marked by only one of the two annotators.
Negation instances are aligned across annotators through their cue tokens.
Proposals whose cue tokens overlap without being identical are merged into one instance (12 cases, typically a multi-word cue that one annotator marked only partially).
In one sentence, the alignment is ambiguous; its separate proposals are retained, flagged and excluded from model fitting.
\paragraph{Scope consolidation.}
The reference contains 342 negation instances.
For 178, both annotators assign the same scope, which is retained.
For 76, only one annotator provides the aligned cue and that annotator's scope is copied.
The remaining 88 instances contain 312 disputed tokens; tokens on which both annotators agree retain their label.
We resolve the disputed tokens with a binary Dawid--Skene model \citep{Dawid:Skene:1979}, which estimates a sensitivity and specificity for each annotator and combines the observed labels into the posterior probability that a token belongs to the scope.
Tokens with a posterior of at least 0.5 are assigned to the scope; 136 of the 312 disputed tokens have a posterior between 0.4 and 0.6.
Parameters are estimated with MAP-EM on all tokens (agreed and disputed) of training-split instances annotated by both annotators with an unambiguous alignment (2,646 tokens) and then applied unchanged to all splits, so validation and test annotations do not influence them.
We use weak $\mathrm{Beta}$ priors with mode 0.9 on sensitivity and specificity.
The estimated sensitivities are 0.91, 0.92 and 0.98 and the specificities 0.99, 0.98 and 0.94 for A1--A3.
\subsection{Training Details}
\label{app:neg-training}
Our models follow the \texttt{D-Neg} training setup, with four modifications.
First, labels are assigned only to the first subword of each word, ensuring that training and evaluation are performed at the word level.
Second, because the EuroBERT tokenizer removes the leading space from pre-segmented words, we restore it before tokenization so that subword segmentation matches that obtained from running text.
Third, before dependency parsing, we remove punctuation attached to words, preserving a one-to-one correspondence between input words and parser tokens.
Fourth, we train for 10 epochs and select the checkpoint with the highest validation F1 rather than the lowest validation loss.
Training sentences exceeding the maximum sequence length are excluded for the GAT model; no validation or test sentence exceeds this limit.
The released \texttt{D-Neg} models are evaluated using their original inference code without modification.
\subsection{Detailed Results and Analysis}
\label{app:neg-detailed-results}
Table~\ref{tab:neg-results-detail} reports precision, recall, F1 and exact match for all GAT models.
For cue detection, most \texttt{D-Neg} models achieve high precision (up to 0.94) but lower recall (at most 0.76), whereas our model is more balanced (P 0.88, R 0.89).
In both tasks, most models favor precision over recall. The exceptions are the \texttt{DT-Neg} model for cue detection\footnote{Our model favors recall by only 0.01 percentage points; we therefore treat its performance as effectively balanced rather than as a meaningful preference for recall.} and the \texttt{ConanDoyle-neg} model for scope resolution.
The cue model trained on \texttt{DT-Neg} is a clear outlier: it labels frequent words such as \textit{ich}, \textit{ja} and \textit{auch} as cues, resulting in a precision of 0.06.
\begin{table*}[t!]
\centering
\small
\begin{tabular}{lcccccccc}
\toprule
& \multicolumn{4}{c}{Cue} & \multicolumn{4}{c}{Scope} \\
\cmidrule(lr){2-5}\cmidrule(lr){6-9}
Training data & P & R & F1 & EM & P & R & F1 & EM \\
\midrule
BioScope (abstracts) & 0.94 & 0.42 & 0.58 & 0.89 & 0.60 & 0.42 & 0.49 & 0.19 \\
BioScope (full) & 0.75 & 0.16 & 0.26 & 0.85 & 0.51 & 0.25 & 0.34 & 0.08 \\
ConanDoyle-neg & 0.93 & 0.71 & 0.81 & 0.92 & 0.47 & 0.61 & 0.53 & 0.24 \\
DT-Neg & 0.06 & 0.63 & 0.11 & 0.31 & 0.69 & 0.54 & 0.61 & 0.27 \\
SFU Review & 0.93 & 0.74 & 0.82 & 0.92 & 0.50 & 0.26 & 0.34 & 0.11 \\
SOCC & 0.83 & 0.76 & 0.79 & 0.91 & 0.42 & 0.35 & 0.38 & 0.14 \\
\midrule
Ours (3 seeds) & 0.88,{\scriptsize$\pm$0.05}
& 0.89{\scriptsize$\pm$0.02}
& 0.88{\scriptsize$\pm$0.02}
& 0.95{\scriptsize$\pm$0.01}
& 0.81{\scriptsize$\pm$0.05}
& 0.56{\scriptsize$\pm$0.05}
& 0.66{\scriptsize$\pm$0.02}
& 0.42{\scriptsize$\pm$0.04} \\
\bottomrule
\end{tabular}
\caption{Negation results on our test split for GAT models: word-level precision (P), recall (R) and F1 for the positive class, together with exact match (EM), measured as the proportion of exactly correct sentences for cue detection and exactly correct scopes for scope detection. \texttt{D-Neg} models are applied as released; our results are averaged over three seeds (mean $\pm$ standard deviation). Scope detection uses gold cues.}
\label{tab:neg-results-detail}
\end{table*}
\paragraph{Cue errors.}
The test split contains 38 annotated cue tokens, of which 26 are \textit{nicht} (24) or \textit{nie} (2); all models detect almost every occurrence.
Our model recovers 7.7 of these 12 tokens on average, whereas the strongest \texttt{D-Neg} GAT model recovers two.
The most frequent false positives from our model are \textit{noch}, \textit{verschwommen}, \textit{Abzug} and the tag question \textit{ne?}.
\paragraph{Scope errors.}
Table~\ref{tab:scope-errors} classifies each predicted scope as exact, a strict subset of the reference (\emph{short}), a strict superset (\emph{long}) or neither.
Reference scopes contain 5.1 tokens on average.
Our model predicts shorter scopes (3.6 tokens), with strict subsets constituting its most frequent error type.
The best out of domain \texttt{D-Neg} model for scope resolution, namely the \texttt{DT-Neg} GAT variant, similarly predicts shorter scopes than the reference (4.0 tokens), but produces fewer exact matches (27\% vs.\ 42\%).
\begin{table}[t!]
\centering
\small
\begin{tabular}{lc|cccc}
\toprule
Model & Len. & Exact & Short & Long & Other \\
\midrule
Ours GAT & 3.6 & 42 & 32 & 14 & 12 \\
D-NEG GAT, DT-Neg & 4.0 & 27 & 27 & 24 & 22 \\
\bottomrule
\end{tabular}
\caption{Scope error types on the test split. \emph{Len.} denotes mean predicted scope length in tokens; the mean reference scope length is 5.1 tokens.}
\label{tab:scope-errors}
\end{table}
\section{Questionnaire Items} \label{app:survey_items}
Table~\ref{tab:survey-items} lists the 30 questions of the main questionnaire in the order in which they were administered; batteries and split questions yield 52 items, which together with three scripted recording-consent prompts form the 55 interview items of the \texttt{SurveyItem} catalog.
Eleven items were asked only depending on previous answers and the four income follow-ups (\texttt{t520011}--\texttt{t520014}), were never asked.
The German wording of all 131 items, including the interviewee's experience questionnaire (50 items) and the interviewer's protocol (26 items), is part of the release.
Questions are adapted mostly from NEPS; glosses are ours.
\begin{table}[ht]
\centering
\scriptsize
\setlength{\tabcolsep}{3pt}
\begin{tabular}{p{2.0cm}lp{3.6cm}}
\toprule
Section & Variable & Content (English gloss) \\
\midrule
Satisfaction, partnership, family & \texttt{t514001} & Life satisfaction (0--10) \\
 & \texttt{t731112} & Steady partnership \\
 & \texttt{t731111} & Living with partner \\
 & \texttt{t731110} & Marital status \\
 & \texttt{tp1} & Feelings about being single \\
 & \texttt{tp2} & Disagreements in the relationship \\
 & \texttt{t66900a} & Epistemic curiosity (scale) \\
 & \texttt{tn1} & Narcissism and Machiavellianism (scale) \\
Education and employment & \texttt{t731810} & Vocational training or study in the last two years \\
 & \texttt{t731813} & Type of vocational qualification \\
 & \texttt{t731816} & Type of tertiary degree \\
 & \texttt{t731901} & Employment \\
 & \texttt{t731903} & Main activity \\
Earned income & \texttt{t731902} & Weekly working hours \\
 & \texttt{t520010} & Net earned income (open) \\
 & \texttt{t520011} & Net earned income (split) \\
 & \texttt{t520012} & Income bracket below 1,500\,EUR \\
 & \texttt{t520013} & Income bracket 1,500--3,000\,EUR \\
 & \texttt{t520014} & Income bracket above 3,000\,EUR \\
 & \texttt{t731904} & Occupation \\
 & \texttt{ta1} & Dream occupation (open) \\
 & \texttt{ta2} & How education could have helped (open) \\
Locus of control & \texttt{t67010a} & Locus of control (scale) \\
Belonging and trust & \texttt{t517400} & Social belonging \\
 & \texttt{t517100} & Social trust \\
 & \texttt{td1} & Offline activities (battery) \\
 & \texttt{td5} & Online activities (battery) \\
 & \texttt{th18153} & Learning with computer or internet, last 12 months \\
 & \texttt{td9} & Learning programs used (open) \\
 & \texttt{td10} & Preferred survey participation mode (open) \\
\bottomrule
\end{tabular}
\caption{Main questionnaire read out in VR.}
\label{tab:survey-items}
\end{table}

\begin{figure*}[h]
    \centering
    \includegraphics[width=0.22\linewidth]{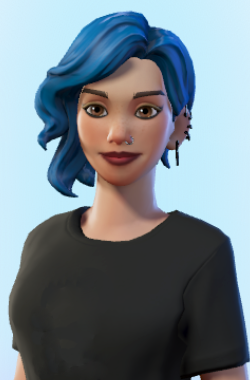}
    \includegraphics[width=0.22\linewidth]{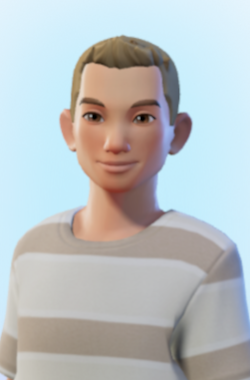}
    \includegraphics[width=0.22\linewidth]{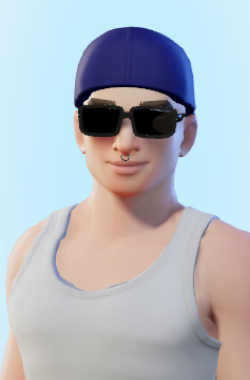}
    \includegraphics[width=0.22\linewidth]{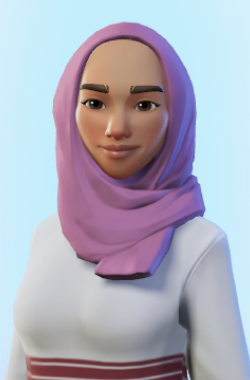}
    \caption{The four avatars used during the interview. From A on the left to D on the right.}
    \label{app:avatars}
\end{figure*}

\begin{table*}[htbp]
\centering
\small
\begin{tabularx}{\textwidth}{llXcr}
\toprule
Layer & Table & Content & Rate & Records \\
\midrule
Session     & \texttt{Experiment}          & interview metadata                          & --        & 27 \\
            & \texttt{Player}              & speaker per interview, connection sessions  & --        & 54 \\
\midrule
Tracking    & \texttt{Eye}                 & binocular gaze poses, validity, confidence  & 18.4\,Hz  & 1,350,572 \\
            & \texttt{Head}, \texttt{Body} & 6-DoF head and rig pose                     & 18.4\,Hz  & 1,350,572 each \\
            & \texttt{Facial}              & 63 blendshape weights                       & 18.4\,Hz  & 1,350,572 \\
            & \texttt{Left/RightHand}      & 6-DoF wrist pose                            & 18.4\,Hz  & 1,350,572 each \\
            & \texttt{Left/RightFinger}    & hand root/pointer pose, pinch, confidence   & 18.4\,Hz  & 515,995 / 544,280 tracked \\
\midrule
Language    & \texttt{Word}                & automatic word tokens, time-aligned         & --        & 53,475 \\
            & \texttt{ReferenceWord}       & post-edited word tokens, forced-aligned     & --        & 55,764 \\
\midrule
Instrument  & \texttt{InterviewItem}       & scripted question stems and battery statements  & --        & 56 \\
            & \texttt{QuestionWindow}      & item readings located by aligning the script to the interviewer transcript   & --        & 1,147 \\
\midrule
Questionnaire & \texttt{SurveyItem}        & items and value labels of three questionnaires (55 interview, 26 protocol, 50 experience) & -- & 131 \\
            & \texttt{SurveyResponse}      & one response set per interview              & --        & 27 \\
            & \texttt{SurveyAnswer}        & answers with missingness codes              & --        & 3,537 \\
\bottomrule
\end{tabularx}
\caption{Released data layers of the 27 interviews. Finger tables hold a row for every frame; the counts give the frames in which the hand was tracked.}
\label{tab:schema}
\end{table*}

\end{document}